\documentclass[letterpaper, 10 pt, conference]{ieeeconf}  % Comment this line out if you need a4paper

\IEEEoverridecommandlockouts                              % This command is only needed if 
\usepackage{graphics} % for pdf, bitmapped graphics files
\usepackage{epsfig} % for postscript graphics files
\usepackage{mathptmx} % assumes new font selection scheme installed
\usepackage{times} % assumes new font selection scheme installed
\usepackage{amsmath} % assumes amsmath package installed
\usepackage{amssymb}  % assumes amsmath package installed
\usepackage{multirow}
\usepackage{graphicx}
\usepackage{diagbox}
\usepackage{hyperref}
\title{\LARGE \bf
MS-rPPG: Multi-spectral State Space Model for \\ Remote Photoplethysmography in Driver Monitoring Systems
}

\author{Jiho Choi$^{1}$ and Sang Jun Lee$^{1,*}$% <-this % stops a space
\thanks{*The corresponding author of this paper.}% <-this % stops a space
\thanks{$^{1}$Jiho Choi and Sang Jun Lee are with the Division of Electronics and Information Engineering, Jeonbuk National University, Republic of Korea.
       Emails: {\tt\small \{jihochoi, sj.lee\}@jbnu.ac.kr}}%.
\thanks{This work was supported by the Institute of Information \& Communications Technology Planning \& Evaluation(IITP)-Innovative Human Resource Development for Local Intellectualization program grant funded by the Korea government(MSIT)(IITP-2025-RS-2024-00439292).}
}

\begin{document}

\maketitle
\thispagestyle{empty}
\pagestyle{empty}
%%%%%%%%%%%%%%%%%%%%%%%%%%%%%%%%%%%%%%%%%%%%%%%%%%%%%%%%%%%%%%%%%%%%%%%%%%%%%%%%
\begin{abstract}
Remote photoplethysmography (rPPG) is a camera-based technique for measuring physiological signals, particularly cardiac activity. From the remotely measured signals, heart rate can be estimated, which is crucial for health monitoring.
In this study, we investigate a driver health monitoring system based on remote heart rate estimation. However, driving environments represent uncontrolled settings where videos are subject to varying illumination conditions and frequent head movements.
% Many rPPG studies have demonstrated that deep learning models such as convolutional neural networks and vision transformers can estimate the pulse signals; however, previous studies were mainly evaluated in controlled settings, limiting their applicability for real-world scenarios such as driver monitoring.
We introduce MS-rPPG, a multi-spectral framework that combines RGB with near-infrared (NIR) face video to alleviate rPPG estimation under challenging driving conditions.
To combine the complementary features from two spectral videos, we propose a cross-spectral linear modulation (CSLM) strategy based on frequency-domain analysis.
Moreover, we introduce MS-Mamba, a novel state space model designed to effectively model long-range temporal dependencies while jointly capturing cross-channel interactions between multi-spectral features.
We collected a real-world dataset called MS-Drive, which was recorded from 50 participants while driving the vehicle.
The proposed method was evaluated on the MR-NIRP Car dataset and MS-Drive datasets.
The experimental results indicate that MS-rPPG shows better robustness and heart rate estimation accuracy than previous methods, highlighting its promise for driver health monitoring.
The codes are available at \href{https://github.com/ziiho08/MS-rPPG}{github.com/ziiho08/MS-rPPG}.
\end{abstract}

%%%%%%%%%%%%%%%%%%%%%%%%%%%%%%%%%%%%%%%%%%%%%%%%%%%%%%%%%%%%%%%%%%%%%%%%%%%%%%%%
\section{INTRODUCTION}
Blood volume pulse (BVP) is a fundamental physiological signal used to derive key indicators of human health, including heart rate (HR) and respiration rate.
Photoplethysmography (PPG) is a widely adopted non-invasive technique that measures BVP by detecting variations in light absorption and reflection caused by blood flow driven by cardiac activity.
However, PPG sensors require direct skin contact, which restricts user movements and limits their suitability for long-term health monitoring.
To overcome these limitations, remote photoplethysmography (rPPG) has emerged as a camera-based alternative that does not require physical contact.
Similar to conventional PPG sensors, rPPG captures subtle variations in light reflected from the facial skin in video recordings, enabling contactless physiological monitoring.
Owing to its non-contact characteristics, rPPG holds significant potential for diverse applications, including healthcare~\cite{hajr2025contactless, zhu2024comparative}, affective computing~\cite{li2024end, yu2021facial}, and driver monitoring systems~\cite{magdalena2018sparseppg, ahmed2025ai}.

\begin{figure}[t]
  \centering
   \includegraphics[width=\linewidth]{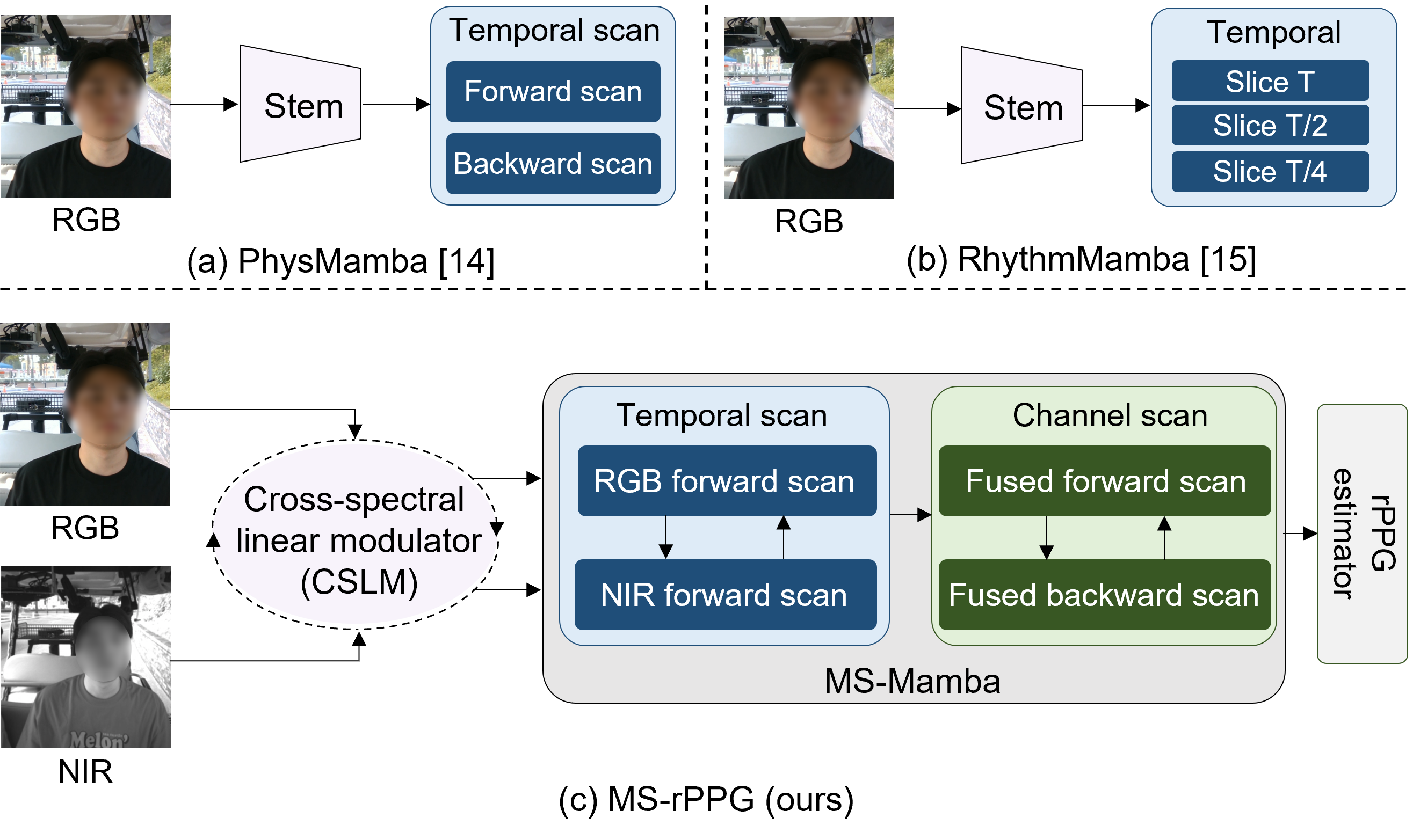}
   \caption{Comparison with recent Mamba-based rPPG frameworks} % 
   \label{fig:intro}
\end{figure}

Early rPPG estimation approaches mainly utilized conventional signal processing methods such as chrominance~\cite{de2013robust} and principal component analysis~\cite{balakrishnan2013detecting}.
Substantial improvements in remote physiological measurement have been achieved with deep learning methods, particularly the convolutional neural network (CNN) and the vision transformer (ViT).
While CNN-based algorithms~\cite{chen2018deepphys, liu2023efficientphys} are effective in extracting spatio-temporal features, they face limitations in capturing long-range temporal dependencies due to their limited receptive fields.
Although ViT-based methods~\cite{yu2022physformer, zou2024rhythmformer} can effectively capture local and global long-range features, the quadratic computational cost of attention restricts their practicality for long video sequences.
These limitations are important to consider when applying rPPG to real-world systems such as driver health monitoring.

Recently, Mamba~\cite{gu2023mamba} has been introduced as a state space model (SSM) that is effective for sequence modeling with reduced computational complexity.
Mamba achieves linear-time complexity while effectively capturing long-term temporal dependencies, facilitating its use in long-sequence tasks such as rPPG estimation.
As shown in Figure 1(a), PhysMamba~\cite{luo2024physmamba} is a lightweight model for real-time rPPG estimation, leveraging linear complexity for efficient long-range temporal modeling.
RhythmMamba~\cite{zou2025rhythmmamba} scans features along the time axis with multiple temporal lengths, enabling effective modeling of both short-term and long-range temporal dependencies in rPPG signals.
However, existing rPPG methods have been evaluated under controlled laboratory conditions, with limited evidence of robustness in unconstrained environments.

This paper focuses on driver heart rate estimation among rPPG applications and addresses the problem of uncontrolled environmental illumination variations.
Varying illumination poses significant challenges for rPPG estimation accuracy, especially for light-susceptible RGB video.
To alleviate the effects of real-world lighting variations, recent studies have explored the use of radio-frequency (RF) signals and near-infrared (NIR) video.
RF signals are robust to lighting variations but remain vulnerable to motion-induced distortions, which are common in dynamic driving scenarios~\cite{vilesov2022blending}.
NIR video is more robust under low-light or rapidly varying illumination, as longer wavelengths are less susceptible to ambient visible light sources~\cite{magdalena2018sparseppg}.
Therefore, we utilize NIR videos along with RGB videos to build a robust model for remote physiological estimation.

As discussed above, this study introduces MS-rPPG, which leverages an SSM with linear computational complexity to capture long-range temporal dependencies from RGB and NIR facial videos in uncontrolled driving environments.
To mitigate drastic illumination changes during driving, we propose a cross-spectral linear modulator (CSLM) based on frequency-domain analysis guided by physiological priors, effectively integrating the complementary information from RGB and NIR images.
Moreover, to enhance long-range temporal dependencies, the multi-spectral Mamba (MS-Mamba) performs two key operations.
First, it conducts a temporal scan across the embeddings for each spectral feature, and it utilizes bidirectional channel scanning to effectively model inter-channel dependencies.

Additionally, real-world driving scenario datasets are limited.
To address this gap, we collected the MS-Drive dataset, which contains RGB and NIR videos with reliable ground truth signals.
We evaluated the proposed framework on both the publicly available MR-NIRP Car dataset and the collected MS-Drive dataset.
Experimental results demonstrate that our approach achieves state-of-the-art performance with HR estimation task under real-world driving conditions, underscoring its potential for deployment in driver monitoring systems.
The main contributions of this paper are:
\begin{itemize}
\item We propose MS-rPPG, a novel multi-spectral framework that integrates RGB and NIR video to enhance the robustness of rPPG estimation in driving environments.
\item We introduce the CSLM based on the frequency domain analysis that effectively fuses the pulse signal from complementary spectral images.
\item We develop MS-Mamba, a state space model designed to capture long-range temporal dependencies with linear computational complexity. MS-mamba scans each spectral video along the temporal axis and performs bidirectional scanning along the channel axis.
\item We collected the real-world driving dataset MS-Drive and conducted extensive experiments on the MR-NIRP Car and MS-Drive datasets, demonstrating that MS-Mamba outperforms previous approaches in terms of remote HR estimation performance under unconstrained driving environments.
\end{itemize}

%%%%%%%%%%%%%%%%%%%%%%%%%%%%%%%%%%%%%%%%%%%%%%%%%%%%%%%%%%%%%%%%%%%%%%%%%%%%%%%%
\section{Related Work}
\subsection{Remote photoplethysmography estimation}
Early rPPG research primarily employed traditional signal processing methods based on hand-crafted features and color space transformations from RGB video~\cite{balakrishnan2013detecting, poh2010non, wang2016algorithmic, de2013robust}.
Recently, deep learning models have established 2D and 3D CNN and ViT-based architectures as the dominant paradigms for rPPG tasks~\cite{yu2019remote, yu2022physformer, zou2024rhythmformer}.
However, relying solely on RGB facial video remains susceptible to external perturbations, including lighting fluctuations and head movement.

To address these limitations, several studies have explored rPPG estimation from NIR images, which exhibit less sensitivity to varying light.
Nowara et al.~\cite{magdalena2018sparseppg} investigated the characteristics of the NIR spectrum, revealing that this spectral band contains minimal energy from sunlight and ambient light. They proposed a robust PCA algorithm and optimized a low-rank matrix to extract the periodic pulse signal, demonstrating effectiveness under varying illumination conditions using the collected real-world MR-NIRP dataset.
However, using only NIR can be challenging because NIR yields a lower signal-to-noise ratio (SNR) than visible light for rPPG due to weaker hemoglobin absorption and reduced sensor sensitivity, leading to much weaker pulsatile signals than RGB.
Park et al.~\cite{park2022self} proposed a self-supervised video-based rPPG framework that integrates RGB and NIR modalities using a dual-branch ViT architecture, achieving enhanced robustness to illumination changes.
Despite using multi-spectral images, ViT inherits substantial computational overhead and has primarily been evaluated under controlled laboratory conditions.

\begin{figure*}[t]
  \centering
   \includegraphics[width=0.95\linewidth]{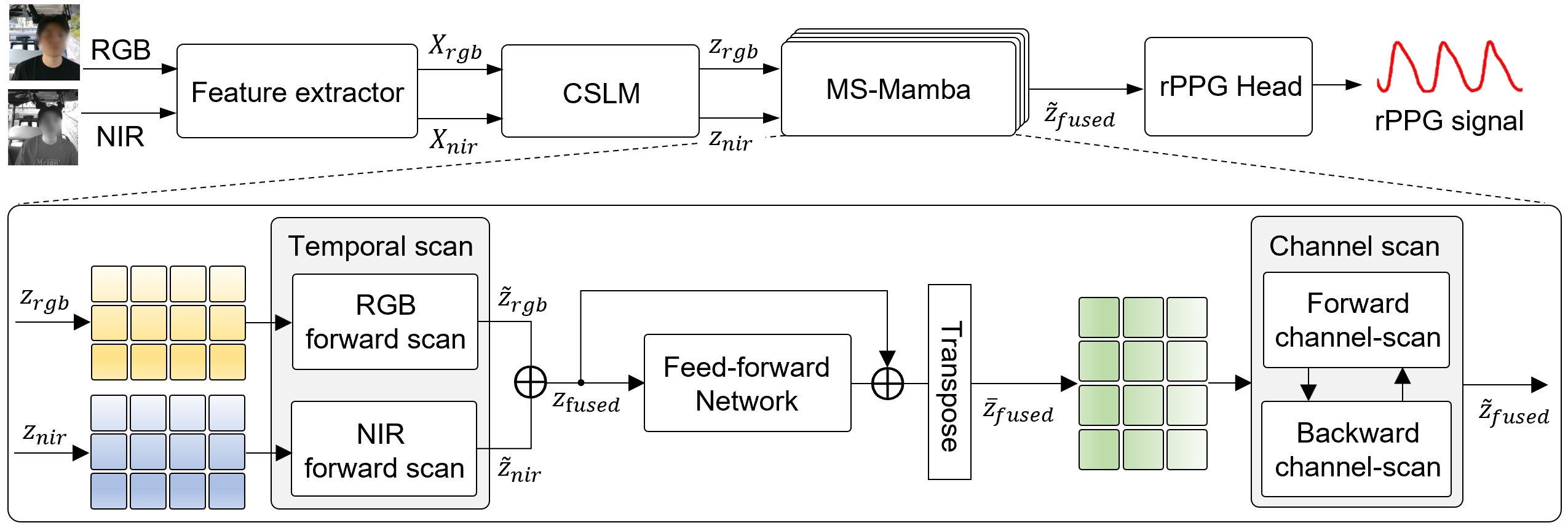}
   \caption{Overview of proposed framework. Our method integrates RGB and NIR face videos for robust remote physiological measurement under real-world driving conditions. MS-rPPG includes the CSLM module for multi-spectral fusion and the MS-Mamba block with temporal and bidirectional channel scanning.}
   \label{fig:proposed}
\end{figure*}

\begin{figure}[ht]
  \centering
   \includegraphics[width=0.8\linewidth]{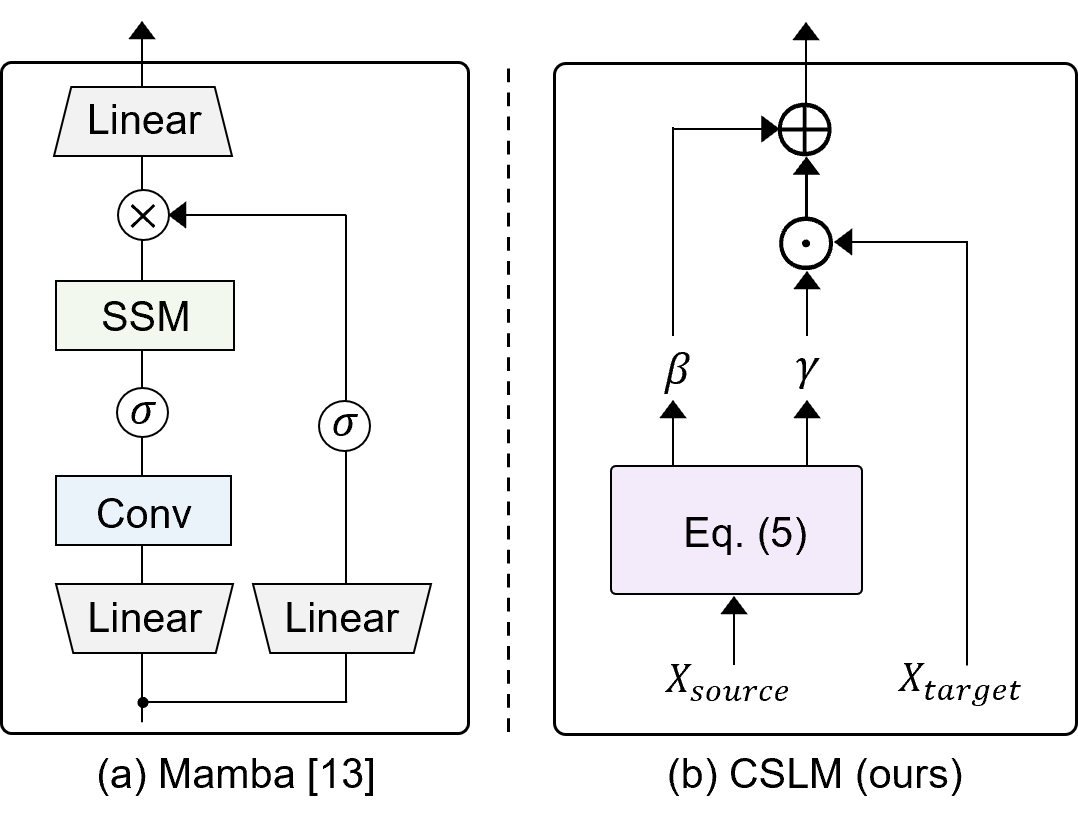}
   \caption{Architecture of the Mamba~\cite{gu2023mamba} and the proposed CSLM.}
   \label{fig:modules}
\end{figure} 

\subsection{State space model in rPPG}
Mamba represents a selective state space model that achieves efficient processing of long-range sequences through linear computational complexity~\cite{gu2023mamba}.
Within computer vision applications, vision Mamba has demonstrated competitive performance across video understanding and image classification tasks, providing an alternative to ViTs with reduced memory requirements and computational latency.

Recent advances in rPPG have increasingly adopted Mamba-based architectures.
PhysMamba~\cite{luo2024physmamba} advances Mamba-based rPPG modeling by introducing a temporal difference Mamba block.
It employs a bidirectional Mamba as depicted in Figure~\ref{fig:intro}(a), achieving superior performance over CNN and Transformer baselines.
As shown in Figure~\ref{fig:intro}(b), RhythmMamba~\cite{zou2025rhythmmamba} established a temporal modeling framework specifically designed for rPPG applications through multi-temporal constraints and frequency-domain feed-forward modules, achieving state-of-the-art performance while preserving computational efficiency.
ME-rPPG~\cite{wang2025memory} further advanced the field by enabling single-frame, low-latency inference with minimal memory consumption, addressing critical real-time deployment requirements.
Unlike existing approaches, our framework models temporal state transitions and simultaneously captures inter-channel dependencies via a bidirectional channel SSM.
This design is an efficient and dynamic channel scanning mechanism.

%%%%%%%%%%%%%%%%%%%%%%%%%%%%%%%%%%%%%%%%%%%%%%%%%%%%%%%%%%%%%%%%%%%%%%%%%%%%%%%%
\section{Method}
\subsection{Preliminaries of Mamba}
To model the sequential physiological data, we adopt the state space model, which describes continuous linear time-invariant (LTI) systems.
An LTI system maps a continuous input signal $x(t) \in \mathbb{R}$ to an output $y(t) \in \mathbb{R}$ through a latent state $h(t) \in \mathbb{R}^N$.
To this end, the system uses a state transition matrix $A \in \mathbb{R}^{N \times N}$, projection matrices $B \in \mathbb{R}^{N \times 1}$ and $C \in \mathbb{R}^{1 \times N}$, and skip connection $D \in \mathbb{R}$.
The system is described by the following ordinary differential equation.
\begin{equation}
\left\{
\begin{array}{l}
h(t)' = A h(t) + B x(t), \\
y(t) = C h(t) + D x(t).
\end{array}
\right.
\end{equation}

To integrate continuous-time SSM into deep learning frameworks, these equations are discretized using the Zero-Order Hold (ZOH) principle.
Specifically, a time scale parameter $\Delta$ is introduced, and the continuous parameters $A$ and $B$ are converted into their discrete counterparts $\bar{A}$ and $\bar{B}$ as follows:
\begin{equation}
    \bar{A} = \exp(\Delta A), \quad 
    \bar{B} = (\Delta A)^{-1} \left( \exp(\Delta A) - I \right) \Delta B,
\end{equation}
where $I$ is the identity matrix of compatible dimensions. 
The discrete state transition and observation equations can then be written as:
\begin{equation}
\left\{
\begin{array}{l}
\displaystyle h_k = \bar{A} h_{k-1} + \bar{B} x_k, \\
y_k = C h_k + D x_k,
\end{array}
\right.
\end{equation}
where $k$ denotes the discrete time step index.

Discrete SSM can be further interpreted as resembling both CNNs and recurrent neural networks.
Specifically, by unrolling the hidden state $h_k$ over the previous time steps and collecting terms of $A$ and $B$, a structured convolutional kernel $\bar{K}$ can be defined.
This enables the following 1D convolution formulation:
\begin{equation}
    \bar{K} = (CB, CAB, \dots, CA^{L-1}B), \quad 
    y = x \ast \bar{K},
\end{equation}
where $\ast$ is a 1D convolution and $L$ is sequence length.
This equation formulation allows for efficient parallelization on modern hardware computation, enabling scalable training and inference for long-sequence modeling tasks.

\subsection{General framework of proposed method}
The proposed framework takes RGB and NIR images as inputs and predicts the rPPG signal.
%RGB $\mathbf{X}_{rgb} \in \mathbb{R}^{T \times 3 \times H \times W}$
The proposed model is composed of three primary components: a feature extractor, CSLM, and MS-Mamba.
First, the feature extractor is composed of 3D convolutions and pooling operations to extract spatio-temporal features, which are crucial for capturing quasi-periodic signals from local spatial regions.
After the feature extractor, we obtain $X_{rgb} \in \mathbb{R}^{T \times C}$ and $X_{nir} \in \mathbb{R}^{T \times C}$, where $T$ and $C$ denotes sequence length and channel dimension, respectively.

Since RGB and NIR signals capture distinct spectral properties, leveraging their respective advantages is essential.
To this end, we introduce the CSLM, which performs feature-wise linear modulation by conditioning one spectral representation on the other.
This process allows each modality to be adaptively refined based on complementary spectral information.
The modulated features are then fed into the MS-Mamba module to model long-term temporal dependencies.
Each modality undergoes sequential modeling along the temporal axis through a state space mechanism, followed by a feed-forward network that enhances frequency-domain representations relevant to physiological rhythms.
We apply a bidirectional channel scan to exploit dependencies across feature channels and further strengthen cross-channel representations.
Lastly, the estimated PPG signal is regressed by the rPPG head module.

\subsection{Frequency domain cross-spectral linear modulation}
The intensity changes in skin regions due to blood flow are subtle compared to the absolute intensity values.
As reported in~\cite{magdalena2018sparseppg}, NIR has a lower SNR than RGB because its lower hemoglobin absorption results in much weaker pulse signals.
While RGB provides stronger periodic signals with a higher SNR in stable lighting conditions, NIR is more beneficial for acquiring physiological signals in low-light environments or under severe illumination changes.
Our goal is to leverage the complementary information between different spectral modalities.
To this end, we introduce CSLM, a fusion method that adopts feature-wise linear modulation~\cite{perez2018film} rather than simple summation or concatenation.
In our approach, CSLM computes scale $\gamma$ and shift $\beta$ parameters from one set of features to adjust and modulate the features from the other modality.

The generator selectively extracts cardiac-related frequency information from an input signal $X_{\text{source}} \in \mathbb{R}^{T \times C}$ to compute the modulation parameters $\gamma$ and $\beta$.
Specifically, the signal is first transformed into the frequency domain via fast Fourier transform (FFT), after which a band-pass mask $M$ is applied to retain only the frequency components corresponding to a predefined physiological range of 40 bpm to 200 bpm.
The filtered signal is then passed through a lightweight neural network $G$, composed of 1D depthwise and pointwise convolutions, which outputs the parameters $\gamma, \beta \in \mathbb{R}^{C \times T}$.
\begin{equation}
\gamma, \beta \;=\; 
G\!\left(F^{-1}\!\big({F}(X_{\text{source}}) \odot M \big) \right),
\end{equation} %\mathcal
where ${F}$ and ${F}^{-1}$ denote the FFT and inverse FFT, respectively, and $\odot$ represents element-wise multiplication.
The generated parameters are constrained to a stable range using the $tanh$ activation function.
Finally, these parameters are used to modulate $X_{\text{target}}$ through an affine transformation, as defined as follows:
\begin{equation}
\text{CSLM}(X_{\text{target}} \mid \gamma, \beta) 
= X_{\text{target}} \odot (1 + \gamma') + \beta',
\end{equation}
where $\gamma'$ and $\beta'$ are the parameters transposed and broadcast to match the dimensions of $X_{\text{target}}$.
The modulated features obtained through this process are denoted as $z_{rgb}$ and $z_{nir}$.
Figure~\ref{fig:modules}(b) depicts the operation of the CSLM.

\subsection{Multi-spectral Mamba}
We propose MS-Mamba, a model for multi-spectral inputs that enhances feature representation by jointly modeling RGB and NIR information.
Mamba has demonstrated strong performance in efficiently capturing long-range dependencies, such as sequential relationships in time-series data.

In our framework, Mamba is leveraged to capture temporal dependencies between the two spectral modalities and facilitate their interaction through a shared state transition matrix.
Initially, the two input streams $z_{rgb}$ and $z_{nir}$ are processed through independent Mamba blocks to learn modality-specific characteristics.
Each block utilizes a shared Mamba layer across both RGB and NIR channels, thereby capturing long-range temporal dependencies while maintaining spectral consistency.
\begin{equation}
\tilde{z}_{m} = \text{Mamba}(z_{m}), \qquad m \in \{\text{rgb}, \text{nir}\}.
\end{equation}
The resulting features, denoted as $\tilde{z}_{rgb}$ and $\tilde{z}_{nir}$, represent the embeddings of each modality.
These features are then fused into $z_{fused}$ and further refined by a frequency-domain feed-forward network (FFN)~\cite{zou2025rhythmmamba}.
FFN is to amplify physiologically relevant periodic components and suppress noise artifacts, yielding clearer pulse representations.

\begin{figure*}[ht]
  \centering
   \includegraphics[width=1.0\linewidth]{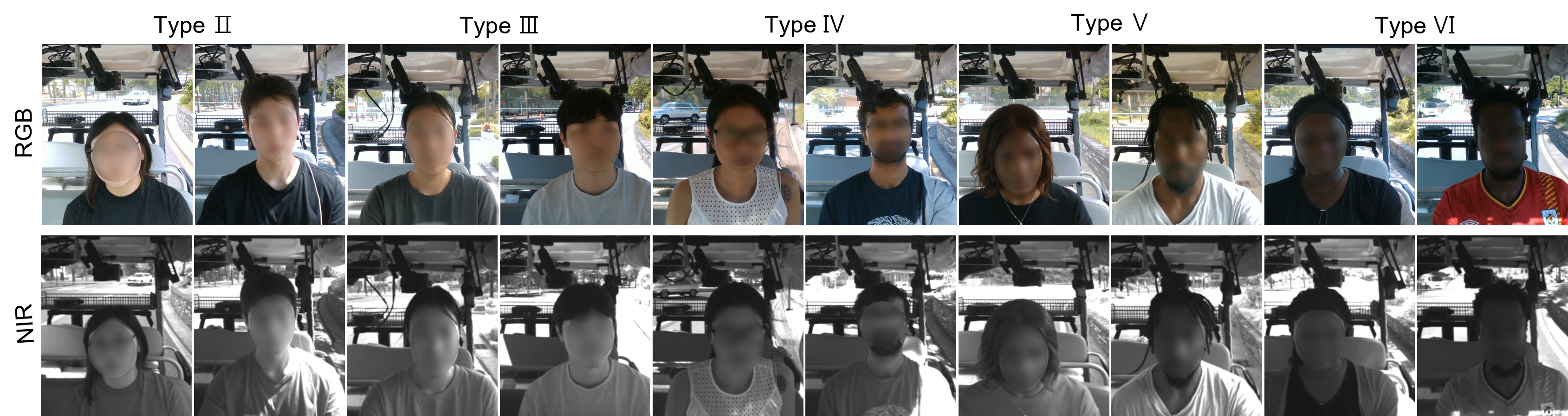}
   \caption{Example images from the MS-Drive dataset, which contains simultaneously captured RGB, NIR, and reliable ECG sensor data.}
   \label{fig:MS-Drive}
\end{figure*}

While Mamba effectively models long-range temporal dependencies, its architecture applies the SSM independently to each channel, overlooking cross-channel interactions and thus limiting representational capacity.
To address this limitation, we reformulate the channel dimension as a sequence by transposing the fused representation $z_{fused}$ to $\bar{z}_{fused} \in \mathbb{R}^{C \times T}$.
The transposition allows the state space model to operate along the channel axis, enabling bidirectional scanning across feature channels.
Consequently, the model effectively captures inter-channel dependencies missed when channels are treated independently, enriching the fused representation with complementary spectral information.
Bidirectional state-space modeling involves both forward and backward scans, as illustrated in Figure~\ref{fig:proposed}.
This operation highlights informative channels and simultaneously models their interactions with other feature channels.
\begin{equation}
\tilde{z}_{fused} = {BiMamba}\!\left(\bar{z}_{fused}\right).
\end{equation}

The output of the channel scanning, denoted as $\tilde{z}_{\text{fused}}$, is added to the $z_{rgb}$ and $z_{nir}$ through skip connections, ensuring that each stream retains its unique modality-specific information while also incorporating the enriched contextual representation obtained from MS-Mamba.
The outputs from the two branches are then averaged to yield the final fused representation, which is fed into the rPPG head to regress the estimated pulse signal.
Our architecture includes 4 sequential MS-Mamba blocks.

\begin{figure}[t]
  \centering
   \includegraphics[width=0.9\linewidth]{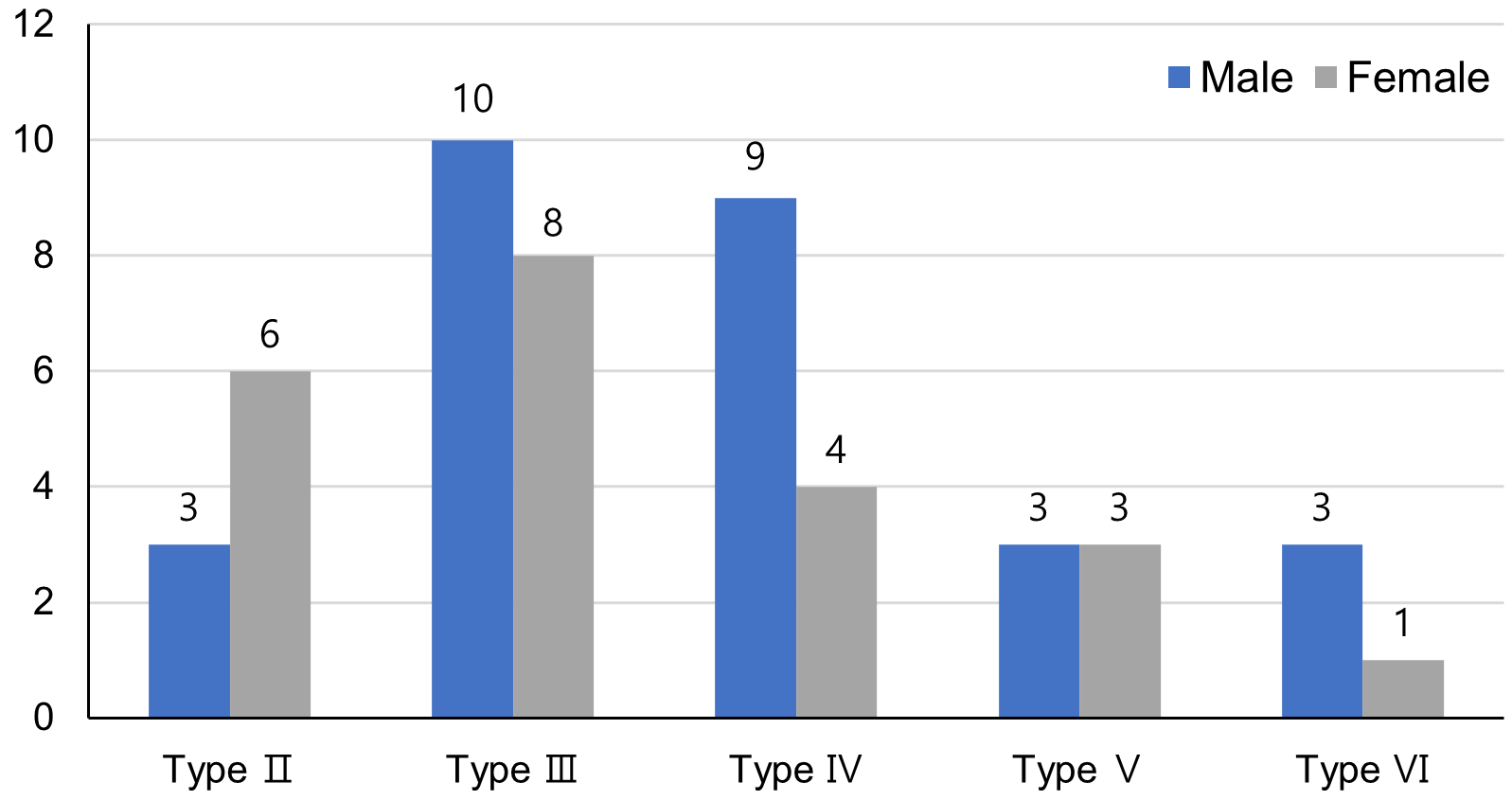}
   \caption{Number of subjects per Fitzpatrick skin type in the MS-Drive dataset.}
   \label{fig:Fitzpatrick}
\end{figure}

\subsection{Loss function}
We employed the negative Pearson correlation loss, a temporal constraint calculated between the estimated rPPG and ground truth signals.
\begin{equation}
   L_{time} =  1- \frac{T\sum^T_1yy'-\sum^T_1y\sum^T_1y'}{\sqrt{(T\sum^T_1y^2-(\sum^T_1y)^2)(T\sum^T_1y'^2-(\sum^T_1y')^2)}},
   \end{equation}
where $y$ and $y'$ are the ground truth pulse signal and predicted rPPG signal, respectively.
$L_{time}$ aligns the trend and peak values of the predicted signal with those of the sensor signal.

For the frequency domain constraint, the loss $L_{freq}$ is defined using cross-entropy.
It is calculated between the heart rate derived from the ground truth signal $y$ and the power spectral density (PSD) of the predicted signal $y'$, as formulated below:
\begin{equation}
L_{freq} = CE(maxIdx(PSD(y)), PSD(y')),    
\end{equation}
where $maxIdx$ represents the index of the maximum value.
The overall loss function is $\alpha\cdot L_{time} + \beta \cdot L_{freq}$,
where $\alpha$ and $\beta$ are set to 0.2 and 1.0, respectively.

\begin{table*}[t]
\centering
\caption{Intra-dataset evaluation results on the MR-NIRP Car dataset with the 975 nm NIR setting under small and large motion scenarios. Best results are marked in bold.}
\addtolength{\tabcolsep}{-0.2em}
\renewcommand{\arraystretch}{1.2}
\resizebox{\textwidth}{!}{%
\begin{tabular}{lcccccccccc}                                                                                                                                       
\hline
\multirow{2}{*}{Methods}                     & \multicolumn{5}{c|}{Small motion}                                 & \multicolumn{5}{c}{Large motion}                                                                           \\ \cline{2-11} 
                         & MAE$\downarrow$   & RMSE$\downarrow$  & MAPE$\downarrow$  & Pearson$\uparrow$  & \multicolumn{1}{c|}{SNR$\uparrow$}   & MAE$\downarrow$      & RMSE$\downarrow$     & MAPE$\downarrow$    & Pearson$\uparrow$   & SNR$\uparrow$               \\ \hline
DeepPhys~\cite{chen2018deepphys}                 & 11.746 & 13.165 & 16.753 & -0.621        & \multicolumn{1}{c|}{-11.486} &  16.336   & 17.443  & 22.675    & -0.003    & -11.715         \\
EfficientPhys~\cite{liu2023efficientphys}       & 12.016 & 15.448 & 15.665 & -0.464        & \multicolumn{1}{c|}{-11.581} &  13.771   & 16.915  & 18.377    & 0.043     & -10.091           \\
PhysNet~\cite{yu2019remote}                      & 6.616  & 10.257 & 10.123 & -0.118        & \multicolumn{1}{c|}{-5.329}  &  6.211    & 7.817   & 9.013     & 0.006     & -5.512             \\
PhysFormer~\cite{yu2022physformer}               & 5.363  & 7.671  & 7.473  & 0.464         & \multicolumn{1}{c|}{-10.777} &  34.833   & 35.738  & 49.78     & 0.229     & -12.479              \\
RhythmFormer~\cite{zou2024rhythmformer}          & 6.211  & 9.077  & 9.358  & 0.034         & \multicolumn{1}{c|}{-1.780}  &  7.021    & 9.125   & 9.991     & 0.581     & -7.854              \\
PhysMamba~\cite{luo2024physmamba}                & 8.101  & 10.834 & 11.696 & -0.362        & \multicolumn{1}{c|}{-6.135}  &  26.462   & 27.391  & 36.206    & NaN       & NaN              \\
RhythmMamba~\cite{zou2025rhythmmamba}            & 7.561  & 10.240 & 11.151 & -0.344        & \multicolumn{1}{c|}{-6.181}  & \multicolumn{1}{c}{9.991} & \multicolumn{1}{c}{13.517} & \multicolumn{1}{c}{14.828
} & \multicolumn{1}{c}{0.254} & \multicolumn{1}{c}{-11.088} \\ \hline
MS-rPPG w/o RGB             & 8.100 & 10.437 & 11.648 & -0.397        & \multicolumn{1}{c|}{-4.959} & \multicolumn{1}{c}{6.886} & \multicolumn{1}{c}{8.635} & \multicolumn{1}{c}{9.400} & \multicolumn{1}{c}{-0.058} & \multicolumn{1}{c}{-5.686} \\
MS-rPPG w/o NIR             & 4.590  & 7.656  & 7.152  & 0.483         & \multicolumn{1}{c|}{-3.668}  & \multicolumn{1}{c}{5.671} & \multicolumn{1}{c}{9.132} & \multicolumn{1}{c}{8.981} & \multicolumn{1}{c}{0.054} & \multicolumn{1}{c}{\textbf{-4.858}} \\
MS-rPPG (ours)                     & \textbf{4.320}  & \textbf{6.038}  & \textbf{6.470}  & \textbf{0.728}         & \multicolumn{1}{c|}{\textbf{-1.606}}  & \multicolumn{1}{c}{\textbf{4.320}} & \multicolumn{1}{c}{\textbf{6.447}} & \multicolumn{1}{c}{\textbf{6.546}} & \multicolumn{1}{c}{\textbf{0.589}} & \multicolumn{1}{c}{-5.228} \\ \hline
\end{tabular}%
}
\label{tab:Intra}
\end{table*}

\begin{figure}[t]
  \centering
   \includegraphics[width=1.0\linewidth]{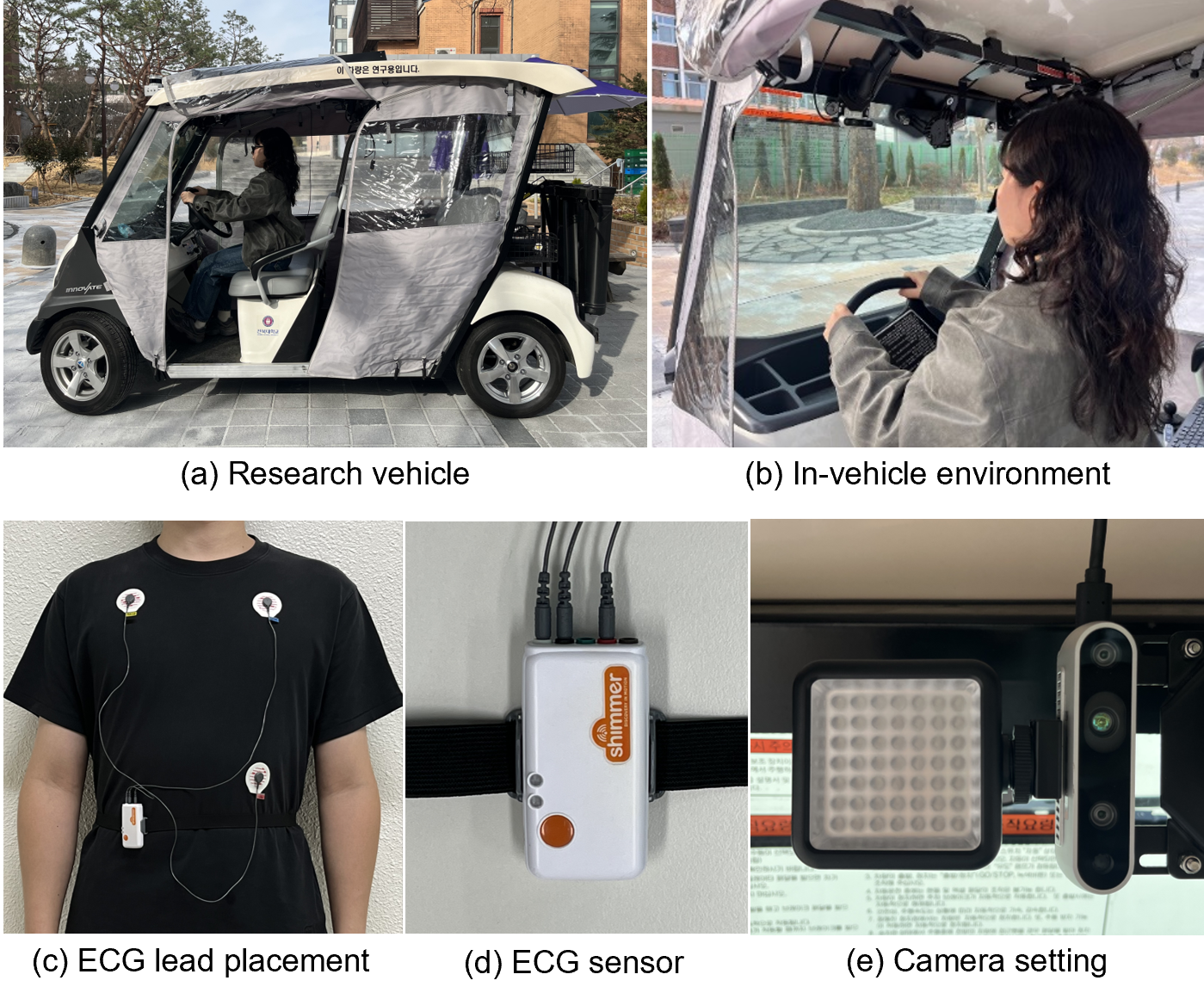}
   \caption{The data acquisition setup for the MS-Drive dataset collection.}
   \label{fig:MS-Drive_set}
\end{figure}

\begin{figure*}[t]
 \centering
  \includegraphics[width=1.0\linewidth]{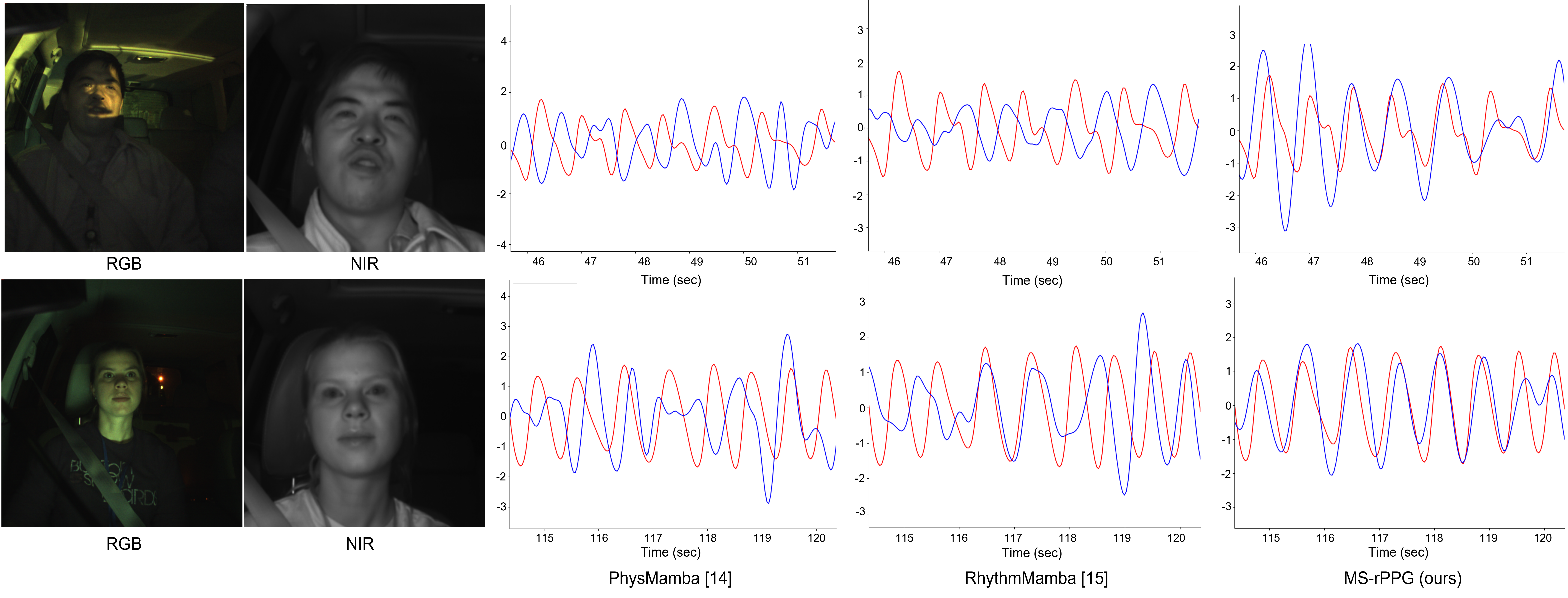}
  \caption{Visualization of test results on the MR-NIRP dataset, showing the ground-truth and predicted PPG signals in red and blue, respectively.}
  \label{fig:rppg}
\end{figure*}

\section{Experimental results}
\subsection{Datasets}
The MR-NIR-Car dataset is a publicly available rPPG dataset that contains facial videos and ground truth pulse signals collected during driving.
The experiments were conducted in both driving and parked conditions inside a garage, and videos were simultaneously recorded using RGB and narrow-band NIR cameras from 18 subjects at 30 FPS with a resolution of 640$\times$640.
The NIR video was captured using narrow-band filters at 940 nm and 975 nm.
Participants were seated in the passenger seat and instructed to remain still, move their heads naturally, and occasionally perform large movements.
The PPG signals were collected using a CMS 50D+ finger pulse oximeter recording at 60 FPS, which was then downsampled to match the video frame rate and synchronized with the video recordings.
Small motion refers to natural head movements that drivers typically make, such as glancing at rear-view and side-view mirrors or talking, without large or sudden movements, while the large motion condition required participants to significantly and abruptly move their heads out of plane.

We present MS-Drive, a real-world driving dataset designed to benchmark remote physiological estimation algorithms.
To ensure diversity in the dataset, we recruited 50 participants (28 male, 22 female) aged 20-34, representing various ethnic backgrounds and distributed across the Fitzpatrick~\cite{fitzpatrick1988validity} skin types.
The distribution is shown in Figure~\ref{fig:Fitzpatrick}.
The dataset includes participants with Fitzpatrick skin types 2-6, accounting for 18\%, 36\%, 26\%, 12\%, and 8\%, respectively. The inclusion of five distinct skin types shows a reasonable diversity in skin tone representation across the dataset.

In contrast to the MR-NIRP Car dataset, participants operated the research vehicle directly during driving sessions.
They were instructed only to avoid looking backward, while all other movements were performed naturally.
For data acquisition, we used an Intel RealSense D435i camera with an external 850 nm NIR illuminator to record synchronized RGB videos at a resolution of $1920\times1080$ and NIR videos at $848\times480$, both at 30 FPS.
Simultaneously, ground truth electrocardiogram (ECG) signals were recorded at 128 Hz using a 3-lead Shimmer3 ECG unit, as shown in Figure~\ref{fig:MS-Drive_set}.
For model training, the physiological signals were synchronized with the video frames.
Example images from the dataset are shown in Figure~\ref{fig:MS-Drive}.
All data collection procedures were conducted under the approval of the Institutional Review Board (IRB Number: JBNU 2025-04-035-002).

\subsection{Implementation Details}
For preprocessing, the RGB and NIR video inputs were first divided into 160 frame segments.
Facial recognition was applied to the first frame of each segment to detect the facial region, which was then cropped and resized to $128\times128$, consistent with existing methods.
We applied the DiffNormalized method~\cite{chen2018deepphys} to input frames, first calculating the normalized difference between consecutive frames.
The resulting sequence is then standardized by its standard deviation.
The proposed method was implemented using an open-source rPPG toolbox~\cite{liu2023rppg} for fair comparisons with existing methods, and the implementation was based on PyTorch.
The models were trained using a batch size of 4 and a learning rate of 3e-4 for 30 epochs.
All experiments were performed with NVIDIA RTX 6000 Ada.
To evaluate the performance of the remote HR estimation task, we employed standard metrics including mean absolute error (MAE), root mean squared error (RMSE), mean absolute percentage error (MAPE), Pearson correlation coefficient ($\rho$), and SNR.

\begin{table*}[t]
\centering
\caption{Cross-dataset evaluation results. We trained the model on MR-NIRP Small Motion 975nm and Large Motion 975nm, and tested on our collected dataset. Best results are marked in bold.}
\resizebox{0.7\textwidth}{!}{%
\addtolength{\tabcolsep}{0.0em}
\renewcommand{\arraystretch}{1.2}
\begin{tabular}{lcccccc}
\hline
\multicolumn{7}{c}{Testing on MS-Drive} \\ \hline
\multirow{2}{*}{Methods} & \multicolumn{3}{c}{Small motion} & \multicolumn{3}{c}{Large motion} \\ \cline{2-7} 
                                                & MAE$\downarrow$ & RMSE$\downarrow$ & MAPE$\downarrow$ & MAE$\downarrow$ & RMSE$\downarrow$  & MAPE$\downarrow$   \\ \hline
%DeepPhys~\cite{chen2018deepphys}                & 16.130	    & 19.399         & 18.339        & 22.970        & 26.338         & 26.285       \\
EfficientPhys~\cite{liu2023efficientphys}       & 25.659        & 28.690         & 29.368        & 31.479        & 33.814         & 36.360       \\
PhysNet~\cite{yu2019remote}                     & 13.189        & 26.397         & 15.073        & 13.345        & 15.939         & 15.449       \\
PhysFormer~\cite{yu2022physformer}              & 14.905        & 17.794         & 17.190        & 22.142        & 25.338         & 29.090        \\
RhythmFormer~\cite{zou2024rhythmformer}         & 13.201        & 15.732         & 15.286        & 11.245        & 14.347         & 13.139        \\
PhysMamba~\cite{luo2024physmamba}               & 14.370        & 17.509         & 16.549        & 38.152        & 39.844         & 44.560         \\
RhythmMamba~\cite{zou2025rhythmmamba}           & 12.373        & 15.867         & 14.273        & 11.785        & 14.257         & 14.414       \\
MS-rPPG (ours)                                            & \textbf{11.797}        & \textbf{14.246}         & \textbf{13.927}      & \textbf{9.601}          & \textbf{12.745}         & \textbf{11.406}        \\ \hline
\end{tabular}%
}
\label{tab:cross_evaluation}
\end{table*}

\subsection{Intra-dataset Evaluation}
Table~\ref{tab:Intra} summarizes the intra-dataset evaluation results on the MR-NIRP dataset under both small- and large-motion conditions using the 975 nm NIR band.
The qualitative results are illustrated in Figure~\ref{fig:rppg}.
Under the small-motion condition, our model achieves an MAE of 4.320 and a Pearson correlation of 0.728, outperforming EfficientPhys with a 7.696 reduction in MAE.
PhysFormer and PhysMamba exhibit unstable performance in the large-motion setting, suggesting that ViT and Mamba-based models are sensitive to such variations.
PhysFormer achieves competitive performance in the small-motion case, with an MAE of 5.363 and an RMSE of 7.671. However, its performance degrades significantly under large head motion, with the MAE and RMSE increasing to 34.833 and 35.748, respectively.
In real-world scenarios, ambient lighting is often distributed unevenly across the face. Under such conditions, significant head movements can induce drastic temporal variations in illumination for the same spatial region, which explains the performance degradation.
MS-Mamba showed significant improvement under large head movements, achieving a 30.513 reduction in MAE and a 0.360 increase in Pearson correlation.

Further analysis highlights the effectiveness of using two spectral modalities, showing that multi-spectral inputs are particularly beneficial under varying lighting conditions.
Removing the RGB input led to a significant drop in rPPG estimation performance.
In contrast, excluding the NIR input resulted in a comparatively minor degradation, especially in the small-motion case.
This suggests that the RGB signal contains high-SNR pulse information, a finding consistent with~\cite{magdalena2018sparseppg}.
While the RGB modality struggled with illumination changes from large head motion, the NIR modality proved more robust under these conditions.
However, surpassing both single-modality approaches, the proposed multi-spectral model consistently achieved superior performance with the lowest errors and highest correlations.
This demonstrates the success of our design in leveraging the complementary advantages of RGB and NIR to achieve consistent improvements over prior methods.

\begin{table}[t]
\centering
\caption{Ablation study of key contributions on the MR-NIRP dataset.}
\renewcommand{\arraystretch}{1.4}
\resizebox{1.0\columnwidth}{!}{%
\addtolength{\tabcolsep}{-0.3em}
\begin{tabular}{ccccccc}
\hline
CSLM         & MS-Mamba       & MAE$\downarrow$   & RMSE$\downarrow$  & MAPE$\downarrow$  & Pearson$\uparrow$ & SNR$\uparrow$    \\ \hline
-                          & \checkmark  & 4.860 & 7.365 & 7.366 & 0.488   & -1.059 \\
1D Conv             & \checkmark  & 4.455 & 6.128 & 6.635 & 0.759   & -2.529 \\ 
\checkmark                       & w/o channel scan & 6.346 & 8.388 & 8.767 & 0.163   & -3.880 \\
\checkmark                       & w/o BiMamba & 6.076 & 7.896 & 8.676 & 0.260   & -2.179 \\
\checkmark                      & \checkmark  & 4.320 & 6.038 & 6.470 & 0.728   & -1.606 \\ \hline
\end{tabular}%
}
\label{tab:ablation1}
\end{table}

\subsection{Cross-dataset Evaluation}
Table~\ref{tab:cross_evaluation} presents the cross-dataset evaluation results, where models trained on the MR-NIRP dataset were tested on MS-Drive.
For the heart rate estimation task, the ground truth heart rate was derived from the collected ECG signal.

Many existing methods, such as EfficientPhys and PhysMamba, struggle to learn robust pulse signal representations, leading to poor generalization in cross-dataset evaluations.
While 3D CNN architecture from PhysNet offered a solid approach for extracting spatio-temporal features, especially under large motion, the MS-rPPG model achieves a promising performance.
Specifically, our proposed method improves upon PhysNet, lowering the MAE by 1.392 in small-motion and 3.744 in large-motion scenarios.

\begin{figure}
 \centering
  \includegraphics[width=1.0\linewidth]{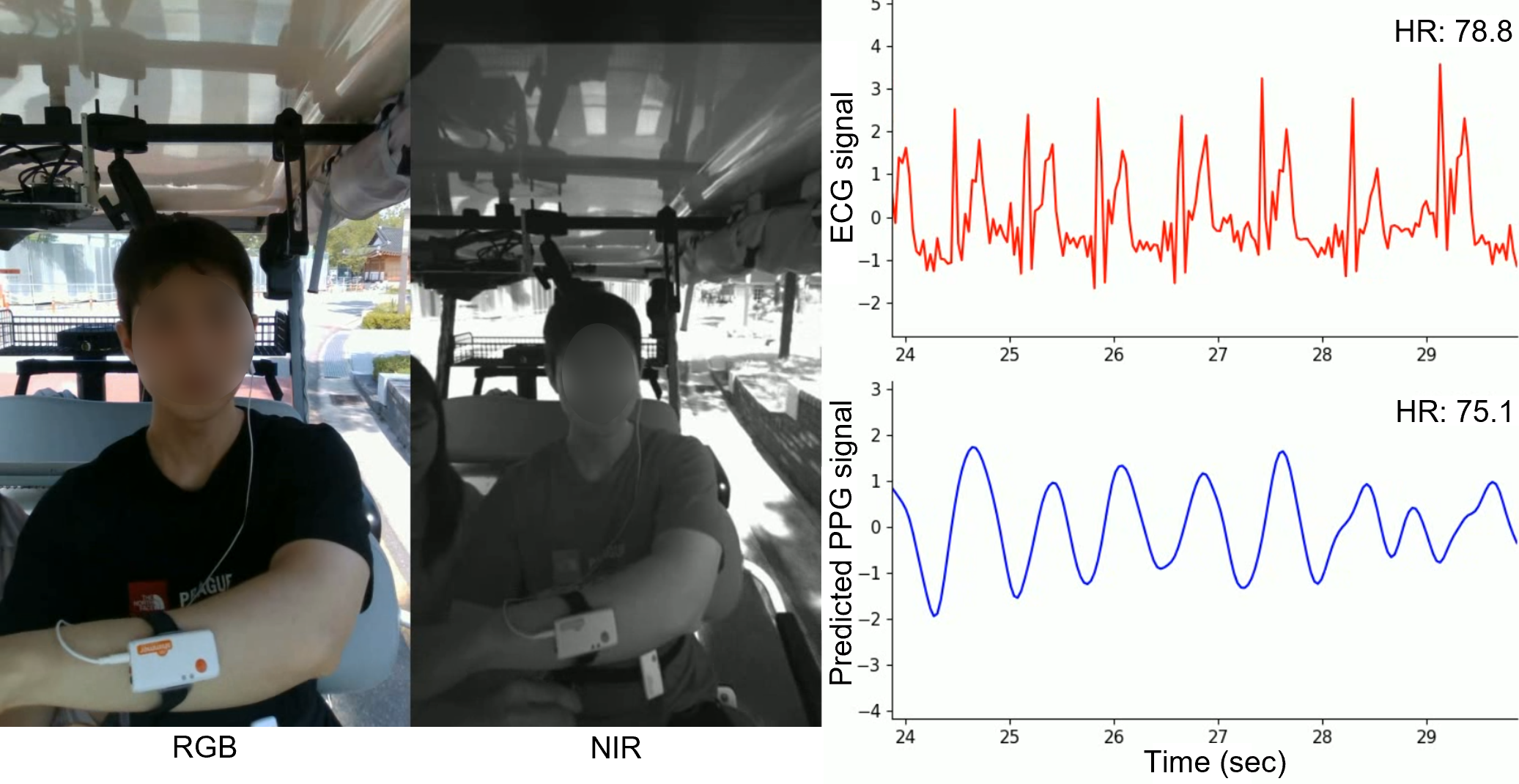}
  \caption{Cross-dataset evaluation results of MS-rPPG on the MS-Drive dataset.}
  \label{fig:cross}
\end{figure}

Although ViT-based models such as RhythmFormer demonstrate strong generalization capability, their quadratic scaling in self-attention is computationally expensive, thereby limiting their practicality for long video sequences.
Our proposed MS-rPPG achieves state-of-the-art performance in cross-dataset rPPG estimation for driving scenarios, with MAE scores of 11.797 under small motion and 9.601 under large motion conditions.
This result represents a notable improvement over recent SSM-based methods such as RhythmMamba, reducing MAE by 0.576 and 2.184, respectively.
Figure~\ref{fig:cross} presents the cross-dataset qualitative results of our model.
These findings confirm the effectiveness of our multi-spectral algorithm for robust rPPG estimation in challenging driving environments.

\subsection{Ablation study}
To demonstrate the effectiveness of the proposed CSLM and MS-Mamba, we conducted an ablation study on small motion 975 nm data from the MR-NIRP dataset. The results are shown in Table~\ref{tab:ablation1}.
We designed CSLM to modulate each feature with complementary spectral information by leveraging physiological frequency band information obtained from FFT-based analysis.
While using a simple 1D convolution for calculating CSLM parameters yielded a high Pearson correlation, it resulted in lower performance across all error metrics and the SNR.
Moreover, when features directly extracted from the feature extractor were fed into MS-Mamba without CSLM, the overall performance declined. However, the drop was not severe, which highlights the inherent robustness of the MS-Mamba architecture.

We also conducted an ablation study on the channel scanning mechanism in MS-Mamba.
The key feature of our proposed framework is its bidirectional channel-axis scan, which is designed to effectively capture inter-channel dependencies.
As shown in the experimental results, unidirectional scanning leads to noticeable performance degradation compared to the proposed method.
This result confirms that bidirectional scanning is crucial for learning robust inter-channel dependencies.
The performance degrades even more significantly when channel scanning is omitted entirely, which further underscores its effectiveness. 
The proposed MS-Mamba and CSLM have proven to be effective for video-based physiological estimation, exhibiting strong robustness against interferences in driving scenarios.

\section{CONCLUSION}
This work proposes MS-rPPG, a novel multi-spectral framework for rPPG estimation in challenging driving environments.
The framework integrates RGB and NIR videos through the CSLM, which adaptively modulates spectral features using a physiological prior derived from FFT analysis.
Additionally, the paper introduces MS-Mamba, a new architecture employing SSM.
MS-Mamba models long-range temporal information with a time-axis SSM and captures inter-channel dependencies via a bidirectional channel SSM.
Comprehensive experiments on the MR-NIRP and MS-Drive datasets demonstrate the effectiveness of the proposed method.
% Intra-dataset evaluations show that the model consistently achieved lower error rates compared to previous methods, while cross-dataset evaluations highlighted its superior generalization to unseen driving scenarios.
% Ablation studies confirm that the FFT-based CSLM substantially improves performance over convolution-based alternatives, and the bidirectional channel scanning mechanism significantly enhances inter-channel modeling.
These results indicate the potential of MS-rPPG for real-world driver monitoring systems, even under challenging conditions with significant motion and illumination variations.

% \addtolength{\textheight}{-12cm}   % This command serves to balance the column lengths
                                  % on the last page of the document manually. It shortens
                                  % the textheight of the last page by a suitable amount.
                                  % This command does not take effect until the next page
                                  % so it should come on the page before the last. Make
                                  % sure that you do not shorten the textheight too much.
% References
\bibliographystyle{IEEEtran} 
\bibliography{ref}            

@inproceedings{balakrishnan2013detecting,
  title={Detecting pulse from head motions in video},
  author={Balakrishnan, Guha and Durand, Fredo and Guttag, John},
  booktitle={Proceedings of the IEEE conference on computer vision and pattern recognition},
  pages={3430--3437},
  year={2013}
}

@article{poh2010non,
  title={Non-contact, automated cardiac pulse measurements using video imaging and blind source separation.},
  author={Poh, Ming-Zher and McDuff, Daniel J and Picard, Rosalind W},
  journal={Optics express},
  volume={18},
  number={10},
  pages={10762--10774},
  year={2010},
  publisher={Optica Publishing Group}
}

@article{wang2016algorithmic,
  title={Algorithmic principles of remote PPG},
  author={Wang, Wenjin and Den Brinker, Albertus C and Stuijk, Sander and De Haan, Gerard},
  journal={IEEE Transactions on Biomedical Engineering},
  volume={64},
  number={7},
  pages={1479--1491},
  year={2016},
  publisher={IEEE}
}

@article{de2013robust,
  title={Robust pulse rate from chrominance-based rPPG},
  author={De Haan, Gerard and Jeanne, Vincent},
  journal={IEEE transactions on biomedical engineering},
  volume={60},
  number={10},
  pages={2878--2886},
  year={2013},
  publisher={IEEE}
}

@article{yu2019remote,
  title={Remote photoplethysmograph signal measurement from facial videos using spatio-temporal networks},
  author={Yu, Zitong and Li, Xiaobai and Zhao, Guoying},
  journal={arXiv preprint arXiv:1905.02419},
  year={2019}
}

@inproceedings{liu2023efficientphys,
  title={Efficientphys: Enabling simple, fast and accurate camera-based cardiac measurement},
  author={Liu, Xin and Hill, Brian and Jiang, Ziheng and Patel, Shwetak and McDuff, Daniel},
  booktitle={Proceedings of the IEEE/CVF winter conference on applications of computer vision},
  pages={5008--5017},
  year={2023}
}

@article{zou2024rhythmformer,
  title={Rhythmformer: Extracting rppg signals based on hierarchical temporal periodic transformer},
  author={Zou, Bochao and Guo, Zizheng and Chen, Jiansheng and Ma, Huimin},
  journal={CoRR},
  year={2024}
}

@inproceedings{yu2022physformer,
  title={Physformer: Facial video-based physiological measurement with temporal difference transformer},
  author={Yu, Zitong and Shen, Yuming and Shi, Jingang and Zhao, Hengshuang and Torr, Philip HS and Zhao, Guoying},
  booktitle={Proceedings of the IEEE/CVF conference on computer vision and pattern recognition},
  pages={4186--4196},
  year={2022}
}

@inproceedings{chen2018deepphys,
  title={Deepphys: Video-based physiological measurement using convolutional attention networks},
  author={Chen, Weixuan and McDuff, Daniel},
  booktitle={Proceedings of the european conference on computer vision (ECCV)},
  pages={349--365},
  year={2018}
}

@inproceedings{magdalena2018sparseppg,
  title={SparsePPG: Towards driver monitoring using camera-based vital signs estimation in near-infrared},
  author={Magdalena Nowara, Ewa and Marks, Tim K and Mansour, Hassan and Veeraraghavan, Ashok},
  booktitle={Proceedings of the IEEE conference on computer vision and pattern recognition workshops},
  pages={1272--1281},
  year={2018}
}

@article{vilesov2022blending,
  title={Blending camera and 77 GHz radar sensing for equitable, robust plethysmography.},
  author={Vilesov, Alexander and Chari, Pradyumna and Armouti, Adnan and Harish, Anirudh Bindiganavale and Kulkarni, Kimaya and Deoghare, Ananya and Jalilian, Laleh and Kadambi, Achuta},
  journal={ACM Trans. Graph.},
  volume={41},
  number={4},
  pages={36--1},
  year={2022}
}

@article{gu2023mamba,
  title={Mamba: Linear-time sequence modeling with selective state spaces},
  author={Gu, Albert and Dao, Tri},
  journal={arXiv preprint arXiv:2312.00752},
  year={2023}
}

@article{park2022self,
  title={Self-supervised RGB-NIR fusion video vision transformer framework for rPPG estimation},
  author={Park, Soyeon and Kim, Bo-Kyeong and Dong, Suh-Yeon},
  journal={IEEE Transactions on Instrumentation and Measurement},
  volume={71},
  pages={1--10},
  year={2022},
  publisher={IEEE}
}

@inproceedings{zou2025rhythmmamba,
  title={RhythmMamba: Fast, Lightweight, and Accurate Remote Physiological Measurement},
  author={Zou, Bochao and Guo, Zizheng and Hu, Xiaocheng and Ma, Huimin},
  booktitle={Proceedings of the AAAI Conference on Artificial Intelligence},
  volume={39},
  number={10},
  pages={11077--11085},
  year={2025}
}

@inproceedings{luo2024physmamba,
  title={PhysMamba: Efficient remote physiological measurement with SlowFast temporal difference mamba},
  author={Luo, Chaoqi and Xie, Yiping and Yu, Zitong},
  booktitle={Chinese Conference on Biometric Recognition},
  pages={248--259},
  year={2024},
  organization={Springer}
}

@article{wang2025memory,
  title={Memory-efficient Low-latency Remote Photoplethysmography through Temporal-Spatial State Space Duality},
  author={Wang, Kegang and Tang, Jiankai and Fan, Yuxuan and Ji, Jiatong and Shi, Yuanchun and Wang, Yuntao},
  journal={arXiv preprint arXiv:2504.01774},
  year={2025}
}

@inproceedings{perez2018film,
  title={Film: Visual reasoning with a general conditioning layer},
  author={Perez, Ethan and Strub, Florian and De Vries, Harm and Dumoulin, Vincent and Courville, Aaron},
  booktitle={Proceedings of the AAAI conference on artificial intelligence},
  volume={32},
  number={1},
  year={2018}
}

@article{liu2023rppg,
  title={rppg-toolbox: Deep remote ppg toolbox},
  author={Liu, Xin and Narayanswamy, Girish and Paruchuri, Akshay and Zhang, Xiaoyu and Tang, Jiankai and Zhang, Yuzhe and Sengupta, Roni and Patel, Shwetak and Wang, Yuntao and McDuff, Daniel},
  journal={Advances in Neural Information Processing Systems},
  volume={36},
  pages={68485--68510},
  year={2023}
}

@article{yu2021facial,
  title={Facial-video-based physiological signal measurement: Recent advances and affective applications},
  author={Yu, Zitong and Li, Xiaobai and Zhao, Guoying},
  journal={IEEE Signal Processing Magazine},
  volume={38},
  number={6},
  pages={50--58},
  year={2021},
  publisher={IEEE}
}

@article{li2024end,
  title={End-to-end multimodal emotion recognition based on facial expressions and remote photoplethysmography signals},
  author={Li, Jixiang and Peng, Jianxin},
  journal={IEEE Journal of Biomedical and Health Informatics},
  year={2024},
  publisher={IEEE}
}

@article{ahmed2025ai,
  title={AI Innovations in rPPG Systems for Driver Monitoring: Comprehensive Systematic Review and Future Prospects},
  author={Ahmed, Soha G and Verbert, Katrien and Siedahmed, Nazar and Khalil, Ashraf and Aljassmi, Hamad and Alnajjar, Fady},
  journal={Ieee Access},
  year={2025},
  publisher={IEEE}
}

@article{zhu2024comparative,
  title={A comparative study of principled rppg-based pulse rate tracking algorithms for fitness activities},
  author={Zhu, Qiang and Wong, Chau-Wai and Lazri, Zachary McBride and Chen, Mingliang and Fu, Chang-Hong and Wu, Min},
  journal={IEEE Transactions on Biomedical Engineering},
  year={2024},
  publisher={IEEE}
}

@article{hajr2025contactless,
  title={Contactless Health Monitoring: An Overview of Video-Based Techniques Utilising Machine/Deep Learning},
  author={Hajr, Alaa and Tarvirdizadeh, Bahram and Alipour, Khalil and Ghamari, Mohammad},
  journal={IET Wireless Sensor Systems},
  volume={15},
  number={1},
  pages={e70009},
  year={2025},
  publisher={Wiley Online Library}
}

@article{fitzpatrick1988validity,
  title={The validity and practicality of sun-reactive skin types I through VI},
  author={Fitzpatrick, Thomas B},
  journal={Archives of dermatology},
  volume={124},
  number={6},
  pages={869--871},
  year={1988},
  publisher={American Medical Association}
}

%\section*{APPENDIX}
%Appendixes should appear before the acknowledgment.

\end{document}